\documentclass[sigconf]{acmart}

\AtBeginDocument{%
  \providecommand\BibTeX{{%
    \normalfont B\kern-0.5em{\scshape i\kern-0.25em b}\kern-0.8em\TeX}}}

\usepackage{latexsym}
\usepackage{graphicx}
\graphicspath{{./images/}}
\usepackage{booktabs} 
\usepackage{color}  
\usepackage{amsmath}  
\usepackage{subcaption}
\usepackage{caption}
\usepackage{tikz}
\usepackage{colortbl} 
\usepackage{framed}
\usepackage{multirow}
\usepackage{multicol}
\usepackage{hyperref}
\usepackage{url}
\usepackage{balance}
\usepackage{verbatim}
\usepackage{cancel}
\usepackage{xspace} 
\usepackage[ruled,vlined]{algorithm2e}
\usepackage{bbold} 
\usepackage{balance}
\usepackage{stmaryrd}
\usepackage{enumitem}
\usepackage{array} 
\usepackage{anyfontsize}

\definecolor{urlcolor}{HTML}{0645AD} 

\newcommand{\struct}[1]{\texttt{\small #1}}
\newcommand{\utterance}[1]{\textit{#1}}
\newcommand{\phrase}[1]{\textit{``#1''}}

\newenvironment{Snugshade}[1][236,236,236]{
    \setlength{\itemsep}{0pt}
     \setlength{\parsep}{0pt}
     \setlength{\topsep}{0pt}
     \setlength{\partopsep}{0pt}
     \setlength{\leftmargin}{1.5em}
     \setlength{\labelwidth}{0em}
     \setlength{\labelsep}{0em} 
    \setlength{\parskip}{0pt}
    \definecolor{shadecolor}{RGB}{#1}%
    \begin{snugshade}
}{%
    \end{snugshade}%
}

\newcommand{\squishlist}{
    \begin{list}{$\bullet$}{ 
        \setlength{\itemsep}{0pt}
        \setlength{\parsep}{1pt}
        \setlength{\topsep}{1pt}
        \setlength{\partopsep}{0pt}
        \setlength{\leftmargin}{1.5em}
        \setlength{\labelwidth}{1em}
        \setlength{\labelsep}{0.5em} 
    } 
}

\newcommand{\squishend}{
  \end{list}  }

\newcommand{\myparagraph}[1]{\noindent \textbf{#1}.}

\newcommand{\method}{\textsc{ReQAP}\xspace}
\newcommand{\reqap}{\textsc{ReQAP}\xspace}
\newcommand{\benchmark}{\textsc{PerQA}\xspace}
\newcommand{\perqa}{\textsc{PerQA}\xspace}

\newcommand{\extract}{\texttt{EXTRACT}\xspace}
\newcommand{\retrieve}{\texttt{RETRIEVE}\xspace}
\newcommand{\operator}[1]{\texttt{\MakeUppercase{#1}}}

\newcommand{\relaxed}[1]{\small (${#1})$}
\newcommand{\relaxedb}[1]{\small ($\mathbf{{#1}}$)}

\newcommand{\cog}{\textsc{CodeGen}\xspace}
\newcommand{\rag}{\textsc{Rag}\xspace}

\makeatother
\copyrightyear{2025} 
\acmYear{2025} 
\acmConference[CIKM '25]{the 34nd ACM International Conference on Information and Knowledge Management}{November 10--14, 2025}{Seoul, Korea}

\begin{document}

\title[The \reqap System for Question Answering over Personal Information]{The \reqap System for Question Answering\\over Personal Information}

\author{Philipp Christmann}
\affiliation{%
  \institution{Max Planck Institute for Informatics\\Saarland Informatics Campus}
  \streetaddress{Saarland Informatics Campus}
  \city{Saarbruecken}
 \country{Germany}
}
\email{pchristm@mpi-inf.mpg.de}

\author{Gerhard Weikum}
\affiliation{%
  \institution{Max Planck Institute for Informatics\\Saarland Informatics Campus}
  \streetaddress{Saarland Informatics Campus}
  \city{Saarbruecken}
    \country{Germany}
}
\email{weikum@mpi-inf.mpg.de}

\setcounter{secnumdepth}{4}

\begin{abstract}
Personal information is abundant on users' devices, from structured data in calendar, shopping records or fitness tools, to unstructured contents in mail and social media posts.
This works presents the \method system that supports users with answers for complex questions that involve filters, joins and aggregation over heterogeneous sources.
The unique trait of \method is that it recursively decomposes questions and incrementally builds an operator tree for execution. 
Both the question interpretation and the individual operators make smart use of light-weight language models, with judicious fine-tuning.
The demo showcases the rich functionality for advanced user questions, and also offers detailed tracking of how the answers are computed by the operators in the execution tree.
Being able to trace answers back to the underlying sources
is vital for human comprehensibility and user trust in the system.
\end{abstract}


\maketitle
\begin{figure} [t]
     \includegraphics[width=\columnwidth]{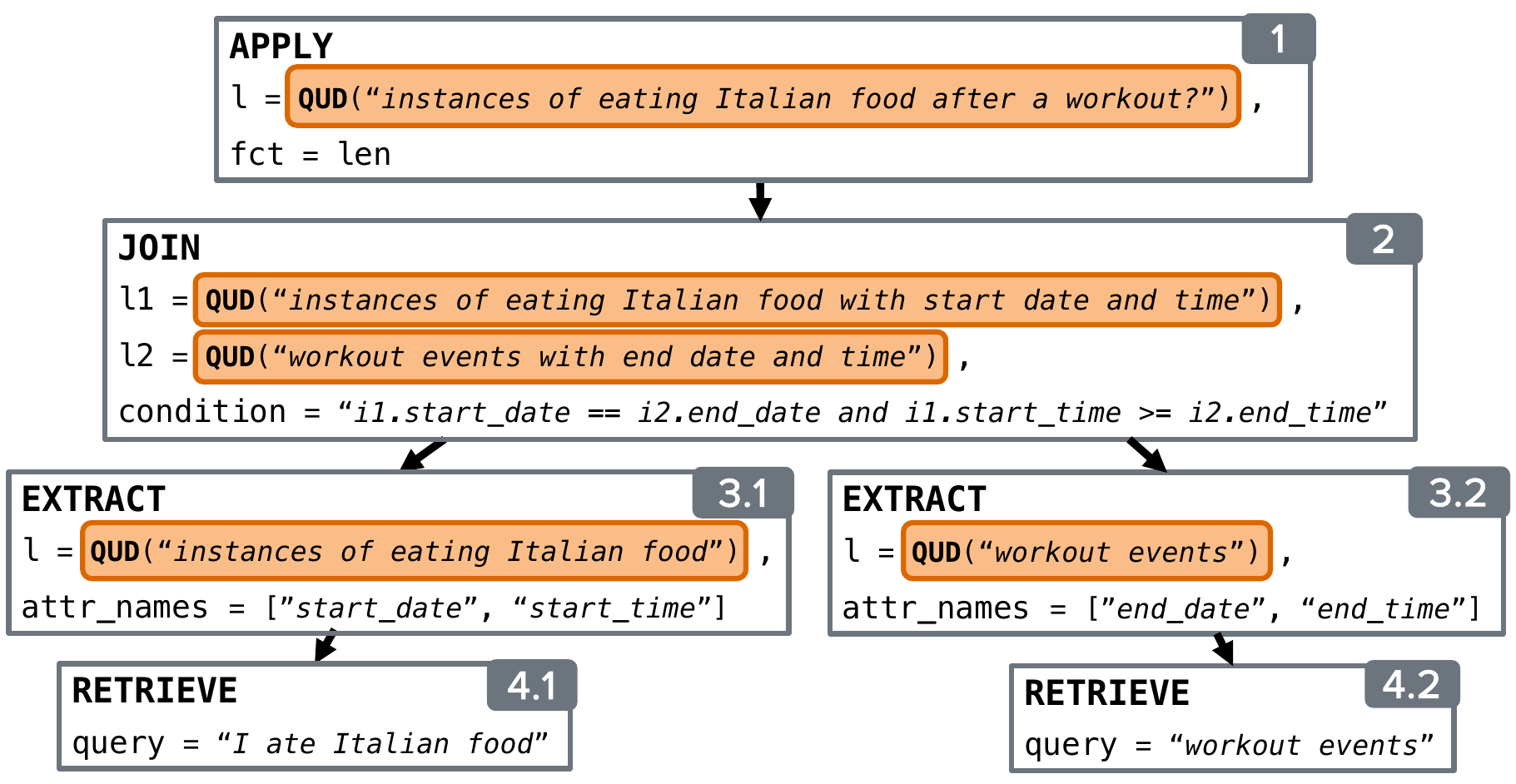}
     \vspace*{-0.7cm}
     \caption{The operator tree constructed by \method for the question \phrase{How often did I eat Italian food after a workout?}.}
     \label{fig:decomposition}
     \vspace*{-0.2cm}
\end{figure}

\vspace*{-0.2cm}
\section{Introduction}
\label{sec:intro}

\myparagraph{Motivation}
Supporting end-users on personal information management involves user-friendly question answering over a diversity of data sources on their mobile devices.
Consider the following example questions:
\begin{Snugshade}
\small
\noindent $q^1$: \utterance{The month I listened to Taylor Swift the most?}\\
$q^2$: \utterance{number of TV series episodes I watched since starting with my thesis?}\\
$q^3$: \utterance{How often did I eat Italian food after a workout?}
\end{Snugshade}
Such questions often require tapping into different sources, some structured and some unstructured:
calendar, fitness, purchases, streaming, mail, social media posts.
For example, $q^3$ needs to aggregate structured data from the fitness app, while having to extract relevant dates from the calendar or notes in the textual body of mails or online posts.
Ideally, question answering (QA) services should run on user devices,
with low footprint (computation, energy, memory) and light-weight language models.
The challenging nature of QA over personal information is further discussed in \cite{halevy2024personal}.


 

\vspace*{0.1cm}
\myparagraph{Approach}
This work presents the \method system,
built upon two novel design concepts:
\squishlist
\item[(i)] {\em recursive decomposition of questions} into simpler sub-questions, 
\item[(ii)] the generation of {\em execution plans} with operators like \operator{Filter}, \operator{Join}, \operator{Group}, and {\em novel operators} {\retrieve} and {\extract} to uniformly cope with structured and unstructured contents.
\squishend

\vspace*{0.1cm}
\noindent For the decomposition, \method employs a fine-tuned large language model (LLM) of moderate size, such as Llama-3.2 (1B).
Unlike established methods for
\textit{code generation},
aka NL2SQL \cite{DBLP:journals/vldb/KatsogiannisMeimarakisK23,DBLP:journals/corr/abs-2408-05109}, \method incrementally builds partial operator trees,
leading to more robust execution plans.
As this compile-time step does not yet access any user data, it could be run as a cloud service,
with the final execution plan transferred to the phone or tablet.

An example for the generated operator trees is shown in Figure~\ref{fig:decomposition}: this is the execution plan for question $q^3$. 
The tree includes a join to combine intermediate results from two sub-questions.
It also shows the novel operators \retrieve and \extract,
which run over both structured and unstructured sources,
treating them as collections of events with
key-value pairs.
The latter are usually mapped to attribute names for structured data,
but involve soft-matching keys against parts of textual contents.
Both of these operators leverage small-scale models,
judiciously fine-tuned for the respective targets.
For details of the \method methodology, our \benchmark dataset for training,
and additional experimental comparisons, see \cite{christmann2025recursive}.


\vspace*{0.1cm}
\noindent The \method system and a short video  can be accessed at:
\squishlist
    \item Demo: \href{https://reqap.mpi-inf.mpg.de/demo}{{\color{urlcolor} https://reqap.mpi-inf.mpg.de/demo}}
    \item Video: \href{https://qa.mpi-inf.mpg.de/reqap/demo-video.mp4}{\color{urlcolor} https://qa.mpi-inf.mpg.de/reqap/demo-video.mp4}
\squishend

\begin{figure} [t]
     \includegraphics[width=\columnwidth]{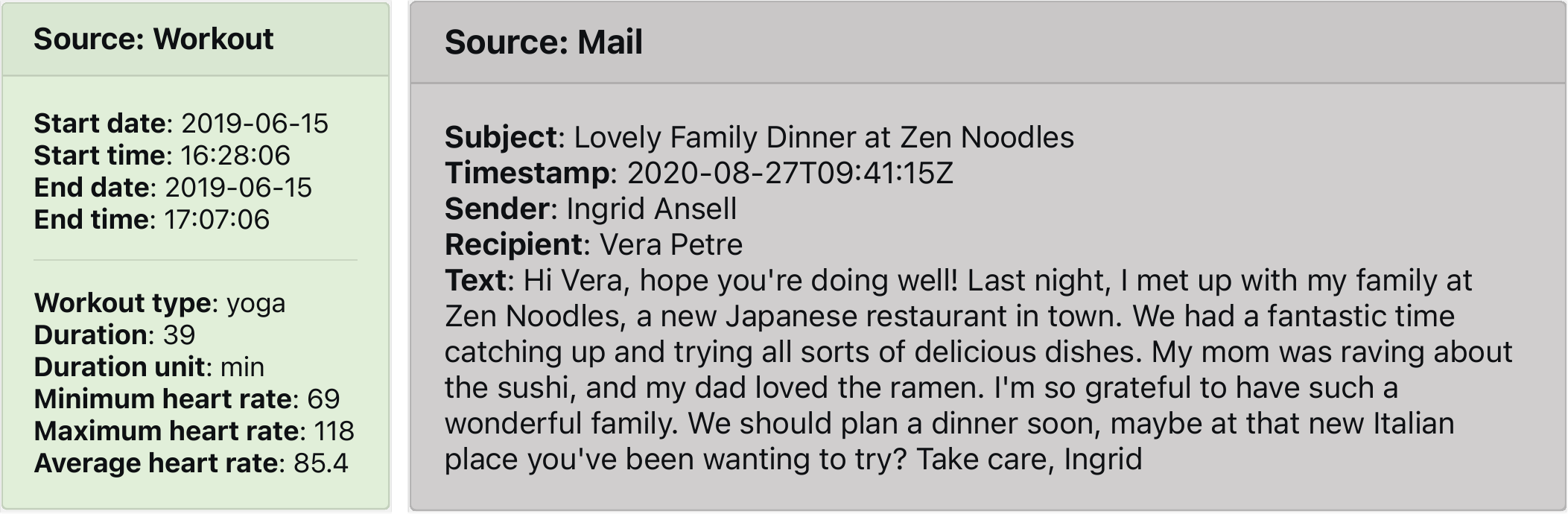}
     \vspace*{-0.7cm}
     \caption{Example workout and mail in the \perqa dataset.}
     \label{fig:user-data}
\end{figure}

\begin{table} [t]
    \small
    \centering
    \newcolumntype{G}{>{\columncolor [gray] {0.90}}c}
    \begin{tabular}{l G c G}
        \toprule
            \textbf{Model $\rightarrow$} & \textbf{GPT4o} & \textbf{LLaMA3.3}  & \textbf{SFT}\\
            \textbf{Method $\downarrow$}  &  ($\gg$100B) &  (70B) & (1B) \\
        \midrule
            \multirow{2}{*}{\textbf{\rag}}    & ${0.149}$  & $0.123$  & $0.029$  \\
                 & \relaxed{0.20} & \relaxed{0.18}  &  \relaxed{0.06} \\

        \midrule
            \multirow{2}{*}{\textbf{\cog}}  & $0.319$ & $0.239$& $0.315$  \\
             &  \relaxed{0.44}  & \relaxed{0.33} & \relaxed{0.47} \\

        \midrule
            \multirow{2}{*}{\textbf{\method} (ours)}    & $\mathbf{0.386}$  & $\mathbf{0.322}$   & $\mathbf{0.380}$   \\
               & \relaxedb{0.52}  &  \relaxedb{0.46}  &  \relaxedb{0.53}  \\
        \bottomrule
    \end{tabular}
    \caption{Experimental results on the \benchmark dataset. Metrics are Hit@1 and, in parentheses, Rlx-Hit@1.}
    \label{tab:main-res}
    \vspace*{-0.6cm}
\end{table}

\begin{table} [t]
    \small
    \center
    \newcolumntype{H}{>{\setbox0=\hbox\bgroup}c<{\egroup}@{}}
    \newcolumntype{G}{>{\columncolor [gray] {0.90}}c}
    \begin{tabular}{l G c c c c}
        \toprule
             & \textbf{QUD $\rightarrow$} & \textbf{XS} & \textbf{S} & \textbf{M} & \textbf{L} \\
             & \textbf{Operators $\downarrow$} & ($135$M) & ($360$M)  &  ($1$B) & ($3$B) \\
        \midrule
            \multirow{2}{*}{\textbf{XS}} &  {\retrieve} (4M)
                & $0.167$ 
                & $0.185$ 
                & $0.193$ 
                & $0.193$ \\ 
            & {\& \extract} (16M)
                & \relaxed{0.28} 
                & \relaxed{0.28} 
                & \relaxed{0.31} 
                & \relaxed{0.30} \\ 
        \midrule
            \multirow{2}{*}{\textbf{S}} &  {\retrieve} (16M) 
                & $0.240$ 
                & $0.287$ 
                & $0.302$ 
                & $0.302$ \\ 
            & {\& \extract} (31M) 
                & \relaxed{0.37} 
                & \relaxed{0.41} 
                & \relaxed{0.42} 
                & \relaxed{0.42} \\ 
        \midrule
            \multirow{2}{*}{\textbf{M}} &  {\retrieve} (23M) 
                & $0.331$ 
                & $0.353$ 
                & $0.378$ 
                & $0.389$ \\ 
           & {\& \extract} (70M)
                & \relaxed{0.48} 
                & \relaxed{0.51} 
                & \relaxed{0.50} 
                & \relaxed{0.52} \\ 
        \midrule
            \multirow{2}{*}{\textbf{L}} & {\retrieve} (33M) 
                & $0.356$ 
                & $0.364$ 
                & $0.396$ 
                & ${0.400}$ \\ 
             & {\& \extract} (139M)
                & \relaxed{0.51} 
                & \relaxed{0.52} 
                & \relaxed{0.54} 
                & \relaxed{0.55} \\ 
        \bottomrule
    \end{tabular}
    \caption{Effect of QUD and operator model sizes on performance.
    Metrics are Hit@1, and Rlx-Hit@1 in parentheses.}
    \label{tab:model-sizes}
    \vspace*{-0.2cm}
\end{table}

\section{The \reqap System}
\label{sec:method}

\reqap answers questions in two stages:
(i) the \textit{question understanding and decomposition (QUD)} that decomposes the question incrementally to construct an operator tree, and
(ii) the \textit{operator tree execution (OTX)} to run the operators and obtain the answer.

Within \reqap, all
personal data is treated as one list of events.
An event can be a music or movie/TV series stream,
a workout, an online purchase, a calendar entry,
a social media post or a mail.
Each event is represented by a set of key-value pairs,
originating from the respective source.
Figure~\ref{fig:user-data}
shows example user data.

\subsection{Question Understanding \& Decomposition}
\label{sec:qud}

The operator tree is constructed incrementally, decomposing 
the complexity of the information need step-by-step.

\vspace*{0.1cm}
\myparagraph{Example}
Consider the operator tree in Figure~\ref{fig:decomposition}:\\
(i) In the first step, the input question is simplified to answering the sub-question \phrase{instances of eating Italian food after a workout?}.\\
An \operator{apply} operator with the length function
as argument is used to derive the answer from the result list.\\
(ii) The second step resolves the temporal condition with a \operator{join},
that combines relevant information for \phrase{eating Italian food\dots}
and \phrase{workout events\dots}.
This leads to two branches with sub-questions.\\
(iii) Each branch (nodes 3.1 and 3.2) is resolved via an
\operator{extract} operation that adds the relevant temporal information.\\
(iv) Finally, the simple sub-questions
\phrase{instances of eating Italian food} and
\phrase{workout events}
are answered via a \retrieve call.

\vspace*{0.1cm}
\myparagraph{Implementation}
We implement the construction of the operator tree
using LLMs. Due to the absence of training data, 
we ran GPT4o on the train set of our \benchmark
benchmark~\cite{christmann2025recursive},
with \textit{in-context learning (ICL)}~\cite{brown2020language},
to derive pairs of questions and operator trees.
Each handcrafted ICL example (we created $40$ in total) consists
of a series of decomposition steps for constructing the operator tree.
The operator trees that lead to the correct answer
are kept for training a smaller scale LLM with $1$ billion parameters.
During inference with the fine-tuned model,
we follow the same decomposition procedure.

\vspace*{-0.1cm}
\subsection{Operator Tree Execution}
\label{sec:otx}

The second stage executes the constructed operator tree, starting
with the leaf nodes (nodes 4.1 and 4.2 in Figure~\ref{fig:decomposition}).

\vspace*{0.1cm}
\myparagraph{\reqap operators}
\reqap integrates the SQL operators \operator{join}, \operator{group\_by} and \operator{unnest},
an operator to filter a result list (\operator{filter}),
operators that enable applying arbitrary functions for
each event (\operator{map}) or a list of events (\operator{apply}),
and standard operators for aggregation (\operator{SUM}, \operator{AVG}, \operator{MAX}, \operator{MIN}, \operator{ARGMAX}, \operator{ARGMIN}).
In addition, we devise two novel operators, \retrieve and \extract,
that enable structured processing of unstructured data.

\vspace*{0.1cm}
\myparagraph{\retrieve operator}
The input to the \retrieve operator is a simple sub-question,
for which relevant events have to be retrieved.
Such events appear in structured and unstructured sources.
For example,
users do not always track workouts with their smart devices,
but also store related calendar entries or post about them
on social media.
Further, entries often match only semantically,
which requires matching beyond surface forms,
and can be in the order of thousands
(e.g., for the query \phrase{music streams})
necessitating a scalable implementation.

To this end, we implement \retrieve as a pipeline:
we apply learned sparse retrieval~\cite{formal2021splade}
to identify candidates, and score this subset of events
using a classifier based on cross-encodings~\cite{DBLP:series/synthesis/2021LinNY}.
For efficiency, we group similar events (e.g., all events from one source)
by identifying patterns of shared content.
Such groups are then scored at once, via a dedicated three-way classifier.
Based on the result, we drop/retain all events or score them independently.
The \retrieve operator also de-duplicates events:
a workout could be covered in a calendar entry
\textit{and} a social media post.
Such duplicates are detected based on
their overlapping
temporal scope.

\vspace*{0.1cm}
\myparagraph{\extract operator}
The \extract operator is used for on-the-fly information extraction:
it processes all given events and adds structured key-value
pairs that can be processed by upstream operators.
This is essential for structured processing of unstructured data.
The relevant attribute names are provided as input.
A simple example is
\struct{date},
which is often
already present in structured form.
More complex examples are
\struct{workout\_type} for a workout mentioned in a social media post,
or \struct{cuisine} for a calendar entry with friends.

For implementing this operator,
we fine-tune a sequence-to-sequence model of small scale,
such as BART~\cite{lewis2020bart} or an efficient $16$M parameter
version of T5~\cite{tay2022scale}, with training data in \benchmark.
The model input is the event and an attribute key for extraction,
and the output is the corresponding value.
The operator returns the input list of events,
augmented with structured key-value pairs.


\subsection{Comparison to the State of the Art}
\label{sec:reqap-expts}
The following presents our key experimental
findings, comparing \reqap against competitive
baselines.
For more detail on our \benchmark dataset,
our configurations, and an in-depth analysis,
refer to \cite{christmann2025recursive}.

\vspace{0.1cm}
\myparagraph{Benchmark}
For training and testing \reqap, we constructed the \benchmark benchmark
with $20$ handcrafted personas and $3{,}500$ complex questions.
For each persona, we created $40{,}000$ events (on average),
utilizing LLM-based verbalization to ensure realistic data.

\vspace{0.1cm}
\myparagraph{Metrics}
We measure \textbf{Hit@1},
and a relaxed version (\textbf{Rlx-Hit@1}) that allows for
deviations of $\pm10\%$ for numeric answers.

\vspace{0.1cm}
\myparagraph{Baselines}
We compare \reqap against baselines for code generation (\textbf{\cog})
or retrieval-augmented generation (\textbf{\rag}).
We provide results based on three different LLMs:
\textbf{GPT4o} ({\small ``{gpt-4o}''}) and \textbf{LLaMA3.3}\footnote{\href{https://huggingface.co/meta-llama/Llama-3.3-70B-Instruct}{\color{urlcolor} meta-llama/Llama-3.3-70B-Instruct}} run with $8$ ICL examples, and a moderate-sized LLM\footnote{\href{https://huggingface.co/meta-llama/Llama-3.2-1B-Instruct}{\color{urlcolor} meta-llama/Llama-3.2-1B-Instruct}} that is fine-tuned for the respective approach (\textbf{SFT}).

\vspace{0.1cm}
\myparagraph{Results}
Table~\ref{tab:main-res} shows the results.
\reqap substantially outperforms baselines in each LLM configuration.
Importantly, our fine-tuned version with $1$B parameters performs
on par with the GPT-based variant (Hit@1) and performs best
when allowing for approximate answers
with a moderate $\pm10\%$ slack (Rlx-Hit@1). 

\vspace{0.1cm}
\myparagraph{Analysis}
Table~\ref{tab:model-sizes}
shows an analysis of the performance of
\reqap for different model sizes
on the \perqa dev set.
We vary the size of the QUD model,
and the models within the \retrieve
and \extract operators.
As can be expected,
there is a substantial gap between the smallest (QUD=XS, operators=XS) and the largest (L, L) variant.
However, there are interesting trade-offs,
and even variants with a very small QUD model
and moderate operate size (XS, M) perform
reasonably well with a Rel-Hit@1 of $48\%$.
Our interactive demo makes use of the (M, L) variant.



\begin{figure} 
     \includegraphics[width=\columnwidth]{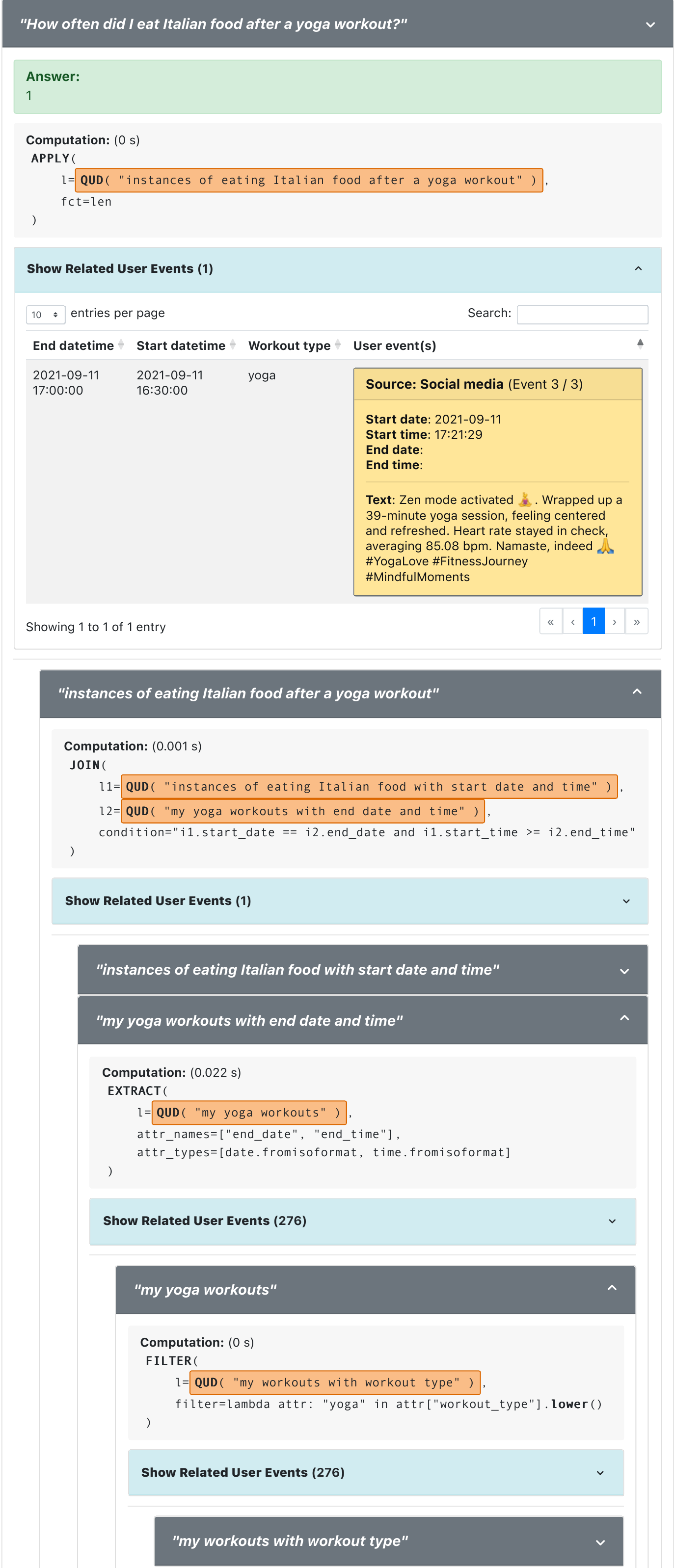}
     \vspace*{-0.5cm}
     \caption{
        Screenshot of the \reqap interface
        for the question \phrase{How often did I eat Italian food after a yoga workout?},
        showing the decomposition steps, operators and related events,
        for tracing the answer derivation.
    }
     \label{fig:demo-tree}
\end{figure}

\begin{figure} [t]
     \includegraphics[width=0.7\columnwidth]{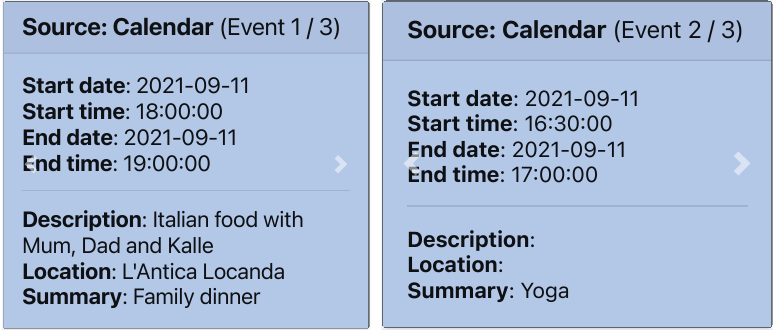}
     \vspace*{-0.3cm}
     \caption{
        Relevant user events for the example question
        \mbox{\phrase{How often did I eat Italian food after a yoga workout?}}.}
     \label{fig:demo-events}
     \vspace*{-0.2cm}
\end{figure}

\section{User Interface}
\label{sec:ui}

\subsection{Personal User Data}
\label{sec:ui-data}

\vspace*{0.1cm}
\myparagraph{User data creation}
For the demo, we utilize $5$
personas from the \perqa dataset,
and selected $4{,}000$
user events
to enable browsing and searching
all of them.
These synthetic personas are handcrafted, with 
details on their family,
career, personal milestones and
preferences (e.g., movies, music or traveling).
For running \reqap locally on their own data
(\reqap loads data exports from the respective services),
users can download our
Docker images\footnote{\href{https://github.com/PhilippChr/ReQAP-docker}{\color{urlcolor} https://github.com/PhilippChr/ReQAP-docker}}.

\vspace*{0.1cm}
\myparagraph{User data browser}
The user can browse through their data with
an interface similar to the one
shown for related events
in Figure~\ref{fig:demo-tree}.
Our demo also shows the handcrafted persona information,
for better understanding of the large-scale personal data.
This is not used within the \method algorithm,
as such structured personal information is unlikely
to be readily available in reality.


\vspace*{-0.5cm}
\subsection{\reqap Inference}
\label{sec:ui-data}

\vspace*{0.1cm}
\myparagraph{\reqap configuration}
In our \reqap interface, 
the user can select the sources to use
for answering.
For the sake of this demo,
questions can also be run on data
of various personas.
The default reference date
(\struct{2024-11-25})
considers the last date of any user event
in the \perqa dataset,
to avoid empty answers for temporal
constraints like \phrase{\dots this year}.
The interface also enables configuring
the QUD generation,
which could be disabled 
when facing lay users.

\vspace*{0.1cm}
\myparagraph{Answer tracing}
Figure~\ref{fig:demo-tree} shows an example of how
the answer derivation can be traced by the end user,
for the question \phrase{How often did I eat Italian food after a yoga workout?}.
In each decomposition step, the user can
see the operator, the sub-question(s) and related user events.
For the first operator (\operator{apply}), there is only a single result.
This result consists of three events:
a calendar entry about a visit of an Italian restaurant,
a calendar entry with \struct{summary}=\phrase{yoga},
and the social media post shown in Figure~\ref{fig:demo-tree}.
Figure~\ref{fig:demo-events} displays the other two entries.
The two events on yoga are duplicate entries, discussing the same
real-life event, as identified in our \retrieve operator.
For such join results and duplicated events,
we allow for investigating the entries by clicking
the left/right arrows, similar to a slideshow.
By enabling to browse
through relevant user events,
our interface helps users
verify the answer correctness.

Below these related events, the interface
illustrates the 
decomposition of the question,
with a \operator{join} and an \extract operator,
similar as in Figure~\ref{fig:decomposition}.
For resolving the user condition of only
considering yoga workouts,
\reqap applies a \operator{filter} to prune out other workouts,
based on string matching in the \struct{workout\_type} key of results.
This key (\struct{workout\_type}) is not readily accessible for most sources
(e.g., the calendar and social media entries in Figure~\ref{fig:demo-events}).
Hence, the subsequent operator
resolving \phrase{my workouts with workout type}
applies an \extract operator to add this
structured key and enable the \operator{filter} operation
(not shown in Figure~\ref{fig:decomposition}
due to space limitations).

To avoid overwhelming users,
only the answer, the first operator and its relevant events
are shown initially, and the decomposition steps can be
conveniently accessed by clicking collapsible items.

\vspace*{0.1cm}
\myparagraph{Debugging}
The interface is also
useful for developers,
as it enables viewing  
both correct and failed operator trees.
Further, per-event details for the
\retrieve and \extract operators are
visible.
For the \retrieve operator, the demo shows 
how the event was classified
(via a pattern or via per-event classification),
and the scores obtained by sparse retrieval and
the classifier.
For the \extract operator,
users can explore the extracted key-value pairs
for each event.

\subsection{Implementation}
\label{sec:ui-impl}
The \reqap backend is implemented using Python,
with Flask for hosting API endpoints.
The frontend is also based on Flask,
and makes use of HTML, CSS and JavaScript.
For the general design we leverage
Bootstrap,
and 
DataTables\footnote{\href{https://datatables.net}{\color{urlcolor} https://datatables.net}}
for our searchable tables.

Besides our public code for the
\reqap\footnote{\href{https://reqap.mpi-inf.mpg.de}{\color{urlcolor} https://reqap.mpi-inf.mpg.de}}
methodology
and our online demo\footnote{\href{https://reqap.mpi-inf.mpg.de/demo}{\color{urlcolor} https://reqap.mpi-inf.mpg.de/demo}},
we provide a Docker image and instructions
to process personal data from popular services
for input in \reqap, 
and Docker images for frontend and backend\footnote{\href{https://github.com/PhilippChr/ReQAP-docker}{\color{urlcolor} https://github.com/PhilippChr/ReQAP-docker}}.

\subsection{Usability}
\label{sec:ui-usability}

We utilized a version of the \reqap interface
for a local user study, enabling the users
to answer their questions against their own 
personal data.
Our study had $20$ local undergraduate students
interact with \reqap, each answering $20$ questions
(= total of $400$ judgments).
Users were provided with Docker containers
to run our system on their own devices, this way
eliminating data leakage risks.
We asked users about their
(i) \textit{certainty} of the answer correctness, and 
(ii) their \textit{understanding} of the answer derivation process.
Users provided scores on a scale from $0-10$ in both cases.
We found that users generally understood whether the answer
is correct or not with a high certainty of $8.6$ (out of $10$)
on average.
Their understanding of the individual steps of the answer derivation
was slightly lower with $7.0$ (out of $10$).
\section{Related Work}
\label{sec:rel}

\noindent {\bf Personal data management.}
Initially motivated by tasks
like desktop search 
\cite{DBLP:conf/cidr/DongH05}
or mail/document search~\cite{DBLP:conf/sigir/DumaisCCJSR03},
this theme has been revived with the increasing abundance of personal data on mobile phones and cloud storage~\cite{halevy2024personal}.
Closest to \method is the TimelineQA project \cite{DBLP:conf/acl/TanD0MSYH23},
which 
explored standard \rag and \cog approaches
for this setting.

\vspace{0.1cm}
\noindent {\bf Multi-Source Question Answering.}
The most popular approach is \textit{verbalization}:
all data is transformed into text and treated uniformly as if were a collection of natural language sentences \cite{DBLP:conf/naacl/OguzCKPOSGMY22,DBLP:conf/acl/YinNYR20,badaro2023transformers,christmann2023explainable,DBLP:conf/acl/ZhangSGXLL24}.
This enables the use of retrieval-augmented LLMs,
with question-relevant text snippets and tables retrieved and fed into the LLM inference.
A major alternative is to leverage LLMs
for \textit{translation} of the question into executable code,
often referred to as
NL2SQL 
\cite{DBLP:journals/vldb/KatsogiannisMeimarakisK23,DBLP:journals/corr/abs-2408-05109,DBLP:conf/nips/PourrezaR23,fan2024combining,DBLP:journals/pvldb/LiLCLT24,DBLP:journals/pvldb/GaoWLSQDZ24,DBLP:conf/sigir/GongS24}.
These require {\em schema} information, though, and extensions to tap into text sources are called for.

\method falls in between the verbalization and translation paradigms:
we utilize LLMs for decomposing questions to generate operator trees (similar to translation),
but introduce novel text-centric retrieval and extraction operators that build upon verbalization.

\section{Conclusion}
\label{sec:conc}


The \method system integrates heterogeneous personal data for question answering.
It decomposes questions to generate executable operator trees.
Answers can be traced back to sources, supporting lay users by human-comprehensible explanations and supporting developers for debugging.
In experimental comparisons, ReQAP outperforms
both RAG and SQL-generation baselines, even when using only light-weight LLMs.
We believe that this work on QA over personal information provides insights towards further research.

\newpage
\section*{GenAI Usage Disclosure}
\label{sec:genai}
GenAI was used for creating realistic user data
in the \perqa benchmark.
In our methodology we use language models
for constructing the operator trees,
and for implementing the \retrieve
and \extract operators.
For run-time efficiency and to reduce energy
costs, we utilize small-scale models.
Finally, we compare against LM-based
baselines in experiments.
We did not use GenAI for coding
or the paper writing.

\balance

\bibliographystyle{ACM-Reference-Format}
\bibliography{references}


\begin{thebibliography}{22}


\ifx \showCODEN    \undefined \def \showCODEN     #1{\unskip}     \fi
\ifx \showDOI      \undefined \def \showDOI       #1{#1}\fi
\ifx \showISBNx    \undefined \def \showISBNx     #1{\unskip}     \fi
\ifx \showISBNxiii \undefined \def \showISBNxiii  #1{\unskip}     \fi
\ifx \showISSN     \undefined \def \showISSN      #1{\unskip}     \fi
\ifx \showLCCN     \undefined \def \showLCCN      #1{\unskip}     \fi
\ifx \shownote     \undefined \def \shownote      #1{#1}          \fi
\ifx \showarticletitle \undefined \def \showarticletitle #1{#1}   \fi
\ifx \showURL      \undefined \def \showURL       {\relax}        \fi
\providecommand\bibfield[2]{#2}
\providecommand\bibinfo[2]{#2}
\providecommand\natexlab[1]{#1}
\providecommand\showeprint[2][]{arXiv:#2}

\bibitem[Badaro et~al\mbox{.}(2023)]%
        {badaro2023transformers}
\bibfield{author}{\bibinfo{person}{Gilbert Badaro}, \bibinfo{person}{Mohammed Saeed}, {and} \bibinfo{person}{Paolo Papotti}.} \bibinfo{year}{2023}\natexlab{}.
\newblock \showarticletitle{{Transformers for Tabular Data Representation: A Survey of Models and Applications}}.
\newblock \bibinfo{journal}{\emph{Transactions of the Association for Computational Linguistics}} (\bibinfo{year}{2023}).
\newblock


\bibitem[Brown et~al\mbox{.}(2020)]%
        {brown2020language}
\bibfield{author}{\bibinfo{person}{Tom Brown}, \bibinfo{person}{Benjamin Mann}, \bibinfo{person}{Nick Ryder}, \bibinfo{person}{Melanie Subbiah}, \bibinfo{person}{Jared~D Kaplan}, \bibinfo{person}{Prafulla Dhariwal}, \bibinfo{person}{Arvind Neelakantan}, \bibinfo{person}{Pranav Shyam}, \bibinfo{person}{Girish Sastry}, \bibinfo{person}{Amanda Askell}, {et~al\mbox{.}}} \bibinfo{year}{2020}\natexlab{}.
\newblock \showarticletitle{{Language Models are Few-Shot Learners}}. In \bibinfo{booktitle}{\emph{Advances in Neural Information Processing Systems (NeurIPS 2020)}}.
\newblock


\bibitem[Christmann et~al\mbox{.}(2023)]%
        {christmann2023explainable}
\bibfield{author}{\bibinfo{person}{Philipp Christmann}, \bibinfo{person}{Rishiraj Saha~Roy}, {and} \bibinfo{person}{Gerhard Weikum}.} \bibinfo{year}{2023}\natexlab{}.
\newblock \showarticletitle{{Explainable Conversational Question Answering over Heterogeneous Sources via Iterative Graph Neural Networks}}. In \bibinfo{booktitle}{\emph{Proceedings of the 46th International ACM SIGIR Conference on Research and Development in Information Retrieval ({SIGIR} 2023)}}.
\newblock


\bibitem[Christmann and Weikum(2025)]%
        {christmann2025recursive}
\bibfield{author}{\bibinfo{person}{Philipp Christmann} {and} \bibinfo{person}{Gerhard Weikum}.} \bibinfo{year}{2025}\natexlab{}.
\newblock \showarticletitle{{Recursive Question Understanding for Complex Question Answering over Heterogeneous Personal Data}}. In \bibinfo{booktitle}{\emph{Findings of the Association for Computational Linguistics (ACL 2025)}}.
\newblock


\bibitem[Dong and Halevy(2005)]%
        {DBLP:conf/cidr/DongH05}
\bibfield{author}{\bibinfo{person}{Xin~Luna Dong} {and} \bibinfo{person}{Alon~Y. Halevy}.} \bibinfo{year}{2005}\natexlab{}.
\newblock \showarticletitle{{A Platform for Personal Information Management and Integration}}. In \bibinfo{booktitle}{\emph{Second Biennial Conference on Innovative Data Systems Research ({CIDR} 2005)}}.
\newblock


\bibitem[Dumais et~al\mbox{.}(2003)]%
        {DBLP:conf/sigir/DumaisCCJSR03}
\bibfield{author}{\bibinfo{person}{Susan~T. Dumais}, \bibinfo{person}{Edward Cutrell}, \bibinfo{person}{Jonathan~J. Cadiz}, \bibinfo{person}{Gavin Jancke}, \bibinfo{person}{Raman Sarin}, {and} \bibinfo{person}{Daniel~C. Robbins}.} \bibinfo{year}{2003}\natexlab{}.
\newblock \showarticletitle{{Stuff I've Seen: {A} System for Personal Information Retrieval and Re-Use}}. In \bibinfo{booktitle}{\emph{Proceedings of the 26th Annual International {ACM} {SIGIR} Conference on Research and Development in Information Retrieval ({SIGIR} 2003)}}.
\newblock


\bibitem[Fan et~al\mbox{.}(2024)]%
        {fan2024combining}
\bibfield{author}{\bibinfo{person}{Ju Fan}, \bibinfo{person}{Zihui Gu}, \bibinfo{person}{Songyue Zhang}, \bibinfo{person}{Yuxin Zhang}, \bibinfo{person}{Zui Chen}, \bibinfo{person}{Lei Cao}, \bibinfo{person}{Guoliang Li}, \bibinfo{person}{Samuel Madden}, \bibinfo{person}{Xiaoyong Du}, {and} \bibinfo{person}{Nan Tang}.} \bibinfo{year}{2024}\natexlab{}.
\newblock \showarticletitle{{Combining Small Language Models and Large Language Models for Zero-Shot NL2SQL}}. In \bibinfo{booktitle}{\emph{Proceedings of the VLDB Endowment (VLDB 2024)}}.
\newblock


\bibitem[Formal et~al\mbox{.}(2021)]%
        {formal2021splade}
\bibfield{author}{\bibinfo{person}{Thibault Formal}, \bibinfo{person}{Benjamin Piwowarski}, {and} \bibinfo{person}{St{\'e}phane Clinchant}.} \bibinfo{year}{2021}\natexlab{}.
\newblock \showarticletitle{{SPLADE: Sparse Lexical and Expansion Model for First Stage Ranking}}. In \bibinfo{booktitle}{\emph{Proceedings of the 44th International {ACM} {SIGIR} Conference on Research and Development in Information Retrieval (SIGIR 2021)}}.
\newblock


\bibitem[Gao et~al\mbox{.}(2024)]%
        {DBLP:journals/pvldb/GaoWLSQDZ24}
\bibfield{author}{\bibinfo{person}{Dawei Gao}, \bibinfo{person}{Haibin Wang}, \bibinfo{person}{Yaliang Li}, \bibinfo{person}{Xiuyu Sun}, \bibinfo{person}{Yichen Qian}, \bibinfo{person}{Bolin Ding}, {and} \bibinfo{person}{Jingren Zhou}.} \bibinfo{year}{2024}\natexlab{}.
\newblock \showarticletitle{{Text-to-SQL Empowered by Large Language Models: {A} Benchmark Evaluation}}. In \bibinfo{booktitle}{\emph{Proceedings of the VLDB Endowment (VLDB 2024)}}.
\newblock


\bibitem[Gong and Sun(2024)]%
        {DBLP:conf/sigir/GongS24}
\bibfield{author}{\bibinfo{person}{Zheng Gong} {and} \bibinfo{person}{Ying Sun}.} \bibinfo{year}{2024}\natexlab{}.
\newblock \showarticletitle{{Graph Reasoning Enhanced Language Models for Text-to-SQL}}. In \bibinfo{booktitle}{\emph{Proceedings of the 47th International {ACM} {SIGIR} Conference on Research and Development in Information Retrieval ({SIGIR} 2024)}}.
\newblock


\bibitem[Halevy et~al\mbox{.}(2024)]%
        {halevy2024personal}
\bibfield{author}{\bibinfo{person}{Alon Halevy}, \bibinfo{person}{Yuliang Li}, {and} \bibinfo{person}{Wang-Chiew Tan}.} \bibinfo{year}{2024}\natexlab{}.
\newblock \showarticletitle{{Personal Manifold: Management of Personal Data in the Age of Large Language Models}}. In \bibinfo{booktitle}{\emph{IEEE 40th International Conference on Data Engineering (ICDE 2024)}}.
\newblock


\bibitem[Katsogiannis{-}Meimarakis and Koutrika(2023)]%
        {DBLP:journals/vldb/KatsogiannisMeimarakisK23}
\bibfield{author}{\bibinfo{person}{George Katsogiannis{-}Meimarakis} {and} \bibinfo{person}{Georgia Koutrika}.} \bibinfo{year}{2023}\natexlab{}.
\newblock \showarticletitle{{A Survey on Deep Learning Approaches for Text-to-SQL}}.
\newblock \bibinfo{journal}{\emph{The VLDB Journal}} (\bibinfo{year}{2023}).
\newblock


\bibitem[Lewis et~al\mbox{.}(2020)]%
        {lewis2020bart}
\bibfield{author}{\bibinfo{person}{Mike Lewis}, \bibinfo{person}{Yinhan Liu}, \bibinfo{person}{Naman Goyal}, \bibinfo{person}{Marjan Ghazvininejad}, \bibinfo{person}{Abdelrahman Mohamed}, \bibinfo{person}{Omer Levy}, \bibinfo{person}{Veselin Stoyanov}, {and} \bibinfo{person}{Luke Zettlemoyer}.} \bibinfo{year}{2020}\natexlab{}.
\newblock \showarticletitle{{BART: Denoising Sequence-to-Sequence Pre-training for Natural Language Generation, Translation, and Comprehension}}. In \bibinfo{booktitle}{\emph{Proceedings of the 58th Annual Meeting of the Association for Computational Linguistics (ACL 2020)}}.
\newblock


\bibitem[Li et~al\mbox{.}(2024)]%
        {DBLP:journals/pvldb/LiLCLT24}
\bibfield{author}{\bibinfo{person}{Boyan Li}, \bibinfo{person}{Yuyu Luo}, \bibinfo{person}{Chengliang Chai}, \bibinfo{person}{Guoliang Li}, {and} \bibinfo{person}{Nan Tang}.} \bibinfo{year}{2024}\natexlab{}.
\newblock \showarticletitle{{The Dawn of Natural Language to {SQL:} Are We Fully Ready?}}. In \bibinfo{booktitle}{\emph{Proceedings of the VLDB Endowment (VLDB 2024)}}.
\newblock


\bibitem[Lin et~al\mbox{.}(2021)]%
        {DBLP:series/synthesis/2021LinNY}
\bibfield{author}{\bibinfo{person}{Jimmy Lin}, \bibinfo{person}{Rodrigo~Frassetto Nogueira}, {and} \bibinfo{person}{Andrew Yates}.} \bibinfo{year}{2021}\natexlab{}.
\newblock \bibinfo{booktitle}{\emph{Pretrained Transformers for Text Ranking: {BERT} and Beyond}}.
\newblock \bibinfo{publisher}{Morgan {\&} Claypool Publishers}.
\newblock
\showISBNx{978-3-031-01053-8}


\bibitem[Liu et~al\mbox{.}(2024)]%
        {DBLP:journals/corr/abs-2408-05109}
\bibfield{author}{\bibinfo{person}{Xinyu Liu}, \bibinfo{person}{Shuyu Shen}, \bibinfo{person}{Boyan Li}, \bibinfo{person}{Peixian Ma}, \bibinfo{person}{Runzhi Jiang}, \bibinfo{person}{Yuyu Luo}, \bibinfo{person}{Yuxin Zhang}, \bibinfo{person}{Ju Fan}, \bibinfo{person}{Guoliang Li}, {and} \bibinfo{person}{Nan Tang}.} \bibinfo{year}{2024}\natexlab{}.
\newblock \showarticletitle{{A Survey of NL2SQL with Large Language Models: Where are we, and where are we going?}}
\newblock \bibinfo{journal}{\emph{arXiv}} (\bibinfo{year}{2024}).
\newblock


\bibitem[O\u{g}uz et~al\mbox{.}(2022)]%
        {DBLP:conf/naacl/OguzCKPOSGMY22}
\bibfield{author}{\bibinfo{person}{Barlas O\u{g}uz}, \bibinfo{person}{Xilun Chen}, \bibinfo{person}{Vladimir Karpukhin}, \bibinfo{person}{Stan Peshterliev}, \bibinfo{person}{Dmytro Okhonko}, \bibinfo{person}{Michael~Sejr Schlichtkrull}, \bibinfo{person}{Sonal Gupta}, \bibinfo{person}{Yashar Mehdad}, {and} \bibinfo{person}{Scott Yih}.} \bibinfo{year}{2022}\natexlab{}.
\newblock \showarticletitle{{UniK-QA: Unified Representations of Structured and Unstructured Knowledge for Open-Domain Question Answering}}. In \bibinfo{booktitle}{\emph{Findings of the Association for Computational Linguistics ({NAACL} 2022)}}.
\newblock


\bibitem[Pourreza and Rafiei(2023)]%
        {DBLP:conf/nips/PourrezaR23}
\bibfield{author}{\bibinfo{person}{Mohammadreza Pourreza} {and} \bibinfo{person}{Davood Rafiei}.} \bibinfo{year}{2023}\natexlab{}.
\newblock \showarticletitle{{{DIN-SQL:} Decomposed In-Context Learning of Text-to-SQL with Self-Correction}}. In \bibinfo{booktitle}{\emph{Advances in Neural Information Processing Systems (NeurIPS 2023)}}.
\newblock


\bibitem[Tan et~al\mbox{.}(2023)]%
        {DBLP:conf/acl/TanD0MSYH23}
\bibfield{author}{\bibinfo{person}{Wang{-}Chiew Tan}, \bibinfo{person}{Jane Dwivedi{-}Yu}, \bibinfo{person}{Yuliang Li}, \bibinfo{person}{Lambert Mathias}, \bibinfo{person}{Marzieh Saeidi}, \bibinfo{person}{Jing~Nathan Yan}, {and} \bibinfo{person}{Alon~Y. Halevy}.} \bibinfo{year}{2023}\natexlab{}.
\newblock \showarticletitle{{TimelineQA: {A} Benchmark for Question Answering over Timelines}}. In \bibinfo{booktitle}{\emph{Findings of the Association for Computational Linguistics ({ACL} 2023)}}.
\newblock


\bibitem[Tay et~al\mbox{.}(2022)]%
        {tay2022scale}
\bibfield{author}{\bibinfo{person}{Yi Tay}, \bibinfo{person}{Mostafa Dehghani}, \bibinfo{person}{Jinfeng Rao}, \bibinfo{person}{William Fedus}, \bibinfo{person}{Samira Abnar}, \bibinfo{person}{Hyung~Won Chung}, \bibinfo{person}{Sharan Narang}, \bibinfo{person}{Dani Yogatama}, \bibinfo{person}{Ashish Vaswani}, {and} \bibinfo{person}{Donald Metzler}.} \bibinfo{year}{2022}\natexlab{}.
\newblock \showarticletitle{Scale Efficiently: Insights from Pretraining and Finetuning Transformers}. In \bibinfo{booktitle}{\emph{The Tenth International Conference on Learning Representations ({ICLR} 2022)}}.
\newblock


\bibitem[Yin et~al\mbox{.}(2020)]%
        {DBLP:conf/acl/YinNYR20}
\bibfield{author}{\bibinfo{person}{Pengcheng Yin}, \bibinfo{person}{Graham Neubig}, \bibinfo{person}{Wen{-}tau Yih}, {and} \bibinfo{person}{Sebastian Riedel}.} \bibinfo{year}{2020}\natexlab{}.
\newblock \showarticletitle{{TaBERT: Pretraining for Joint Understanding of Textual and Tabular Data}}. In \bibinfo{booktitle}{\emph{Proceedings of the 58th Annual Meeting of the Association for Computational Linguistics ({ACL} 2020)}}.
\newblock


\bibitem[Zhang et~al\mbox{.}(2024)]%
        {DBLP:conf/acl/ZhangSGXLL24}
\bibfield{author}{\bibinfo{person}{Heidi~C. Zhang}, \bibinfo{person}{Sina~J. Semnani}, \bibinfo{person}{Farhad Ghassemi}, \bibinfo{person}{Jialiang Xu}, \bibinfo{person}{Shicheng Liu}, {and} \bibinfo{person}{Monica~S. Lam}.} \bibinfo{year}{2024}\natexlab{}.
\newblock \showarticletitle{{{SPAGHETTI:} Open-Domain Question Answering from Heterogeneous Data Sources with Retrieval and Semantic Parsing}}. In \bibinfo{booktitle}{\emph{Findings of the Association for Computational Linguistics ({ACL} 2024)}}.
\newblock


\end{thebibliography}

\end{document}